\documentclass[runningheads]{llncs}

\usepackage[final,year=2024,ID=5817]{eccv}
\usepackage{eccvabbrv}

\usepackage{graphicx}
\usepackage{booktabs}
\usepackage{times}
\usepackage{epsfig}
\usepackage{graphicx}
\usepackage{amsmath}
\usepackage{amssymb}
\usepackage{multirow}
\usepackage{cases}
\usepackage{booktabs}% http://ctan.org/pkg/booktabs
\usepackage{color}
\usepackage{cases}
\usepackage{wasysym}
\usepackage{bm}
\usepackage{floatrow}
\usepackage{sidecap}
\newfloatcommand{capbtabbox}{table}[][\FBwidth]

\usepackage[accsupp]{axessibility}  % Improves PDF readability for those with disabilities.

\usepackage[pagebackref,breaklinks,colorlinks,citecolor=eccvblue]{hyperref}
\usepackage{orcidlink}

\begin{document}

\title{Robust Multimodal Learning via Representation Decoupling} 


\author{Shicai Wei \quad\quad\quad  Yang Luo \quad\quad\quad  Yuji Wang  \quad\quad\quad  Chunbo Luo*}

\authorrunning{Wei et al.}


\institute{School of Information and Communication Engineering \\
University of Electronic Science and Technology of China\\
\email {shicaiwei@std.uestc.edu.cn \{c.luo, luoyang\}@uestc.edu.cn }
}

\maketitle

\begin{abstract}

Multimodal learning robust to missing modality has attracted increasing attention due to its practicality. Existing methods tend to address it by learning a common subspace representation for different modality combinations. However, we reveal that they are sub-optimal due to their implicit constraint on intra-class representation. Specifically, the sample with different modalities within the same class will be forced to learn representations in the same direction. This hinders the model from capturing modality-specific information, resulting in insufficient learning. To this end, we propose a novel Decoupled Multimodal Representation Network (DMRNet) to assist robust multimodal learning. Specifically, DMRNet models the input from different modality combinations as a probabilistic distribution instead of a fixed point in the latent space, and samples embeddings from the distribution for the prediction module to calculate the task loss. As a result, the direction constraint from the loss minimization is blocked by the sampled representation. This relaxes the constraint on the inference representation and enables the model to capture the specific information for different modality combinations. Furthermore, we introduce a hard combination regularizer to prevent DMRNet from unbalanced training by guiding it to pay more attention to hard modality combinations. Finally, extensive experiments on multimodal classification and segmentation tasks demonstrate that the proposed DMRNet outperforms the state-of-the-art significantly.

\end{abstract}

\section{Introduction}

Multimodal learning has yielded significant advancements across a wide array of vision tasks, including classification~\cite{mm_cf1,mm_cf2,mm_cf3}, object detection~\cite{mm_detection1,mm_detection2,mm_detection3}, and segmentation~\cite{rgbd_seg1,rgbd_seg2,rgbd_seg3}. Nevertheless, most of these state-of-the-art approaches assume that models are trained and evaluated with the same modality data. In fact, limited by device~\cite{spcical-gan1,device1}, and working condition~\cite{special-hall1,special-hall2}, it is often very costly or even infeasible to collect complete modality data during the inference stage. Consequently, there exists a compelling need to improve the inference robustness of multimodal models for incomplete input.

% \begin{figure}[ht]
% % \setlength{\belowcaptionskip}{-0.3cm}
% \centering
% \includegraphics[width=1.0\columnwidth]{picture/motivation.png} % Reduce the figure size so that it is slightly narrower than the column. Don't use precise values for figure width.This setup will avoid overfull boxes.
% \caption{Comparison of deterministic and probabilistic embeddings for incomplete multimodal learning. Deterministic embedding captures the invariant feature for all modality combinations via a point vector. decoupled multimodal representation gives a distributional estimation instead. This helps preserve more modality-specific information, increasing the intra-modality and inter-modality feature diversity.}
% \label{knowledge-compare}
% \end{figure}

\begin{figure*}[ht]
\centering
\includegraphics[width=1.0\textwidth]{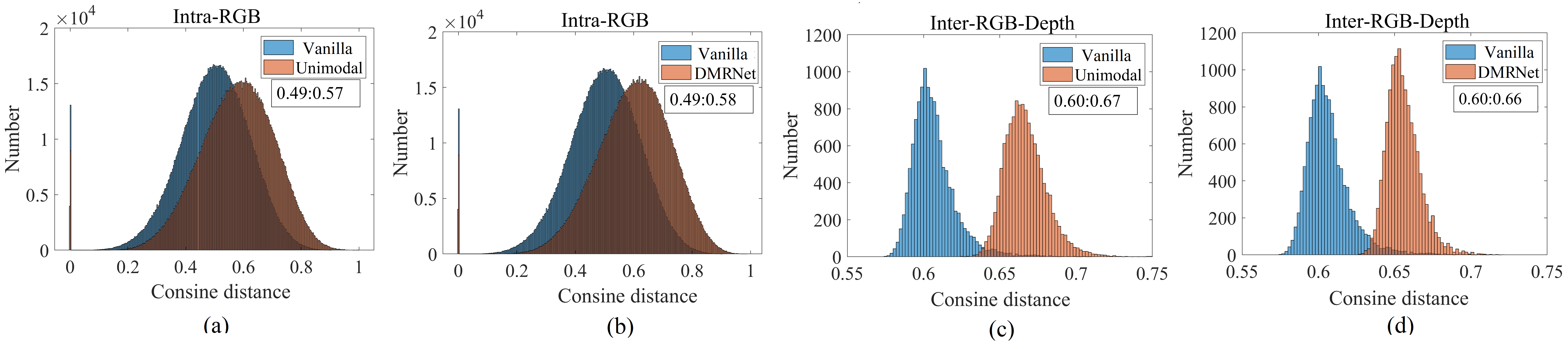} % Reduce the figure size so that it is slightly narrower than the column. Don't use precise values for figure width.This setup will avoid overfull boxes.
\caption{ Illustration of the histogram of the inter-channel distance matrix $D_{channel}$ (defined in Section~\ref{abs}) on the CASIA-SURF dataset. (a) and (b) show the intra-modality inter-channel distance between feature maps of the RGB encoder. (c) and (d) show the inter-modality inter-channel distance between the feature maps of RGB and Depth encoders. `Vanilla' denotes the conventional multimodal model. `Unimodal' means the baseline that trains models for each modality independently, which provides the ideal feature diversity without inter-modality interference. `A:B' denotes the ratio of histogram mean. Higher inter-channel distance means higher diversity .}
\label{diveristy-compare}
\end{figure*}

% Comparison of deterministic and probabilistic embeddings for incomplete multimodal learning. Deterministic embedding captures the invariant feature for all modality combinations via a point vector. decoupled multimodal representation gives a distributional estimation instead. This helps preserve more modality-specific information, increasing the intra-modality and inter-modality feature diversity

 % without considering the modality-specific information

Typically, existing solutions can be divided into two categories: data imputation-based methods and common subspace-based methods. Data imputation-based methods first fill the samples or representations of missing modalities and then apply the traditional multimodal algorithm directly~\cite{special-hall2,special-hall3,special-hall4,special-hall5}. While this type of method is straightforward, it comes with a large computational burden due to the reconstruction of missing modalities, especially when the number of missing modalities increases~\cite{mmformer}. At the same time, the reconstruction may introduce harmful noise and harm the model~\cite{gan-issue1,gan-issue2}. Thus, recent studies mainly focus on the common subspace-based methods that capture the shared feature for all possible input modality combinations to address the incomplete multimodal learning issue~\cite{rfnet,mmformer,lcr,hemis}.

% Although the existing common subspace-based methods are indeed able to increase the efficiency of training and deployment of the multimodal models, their performance is likely to be sub-optimal. Firstly, the variation of input modality combinations is complex, however, the learned embedding needs to be invariant and deterministic. This limits the feature diversity and representation ability, especially when the modality number is large. Secondly, recent methods usually introduce the auxiliary regularizer to force the model to pay more attention to the unimodal input, avoiding overfitting the input with multiple modalities~\cite{mmformer,rfnet}. This assumes that multimodal input is more discriminative than unimodal one. However, our experiments find that the unimodal input could be more discriminative than multimodal input when more than two modalities are considered. Thus existing regularize for unimodal input is also sub-optimal.

Although the existing common subspace-based methods are indeed able to increase the efficiency of training and deployment of the multimodal models, they project the input with different modality combinations into a deterministic embedding and conduct prediction from it directly. This introduces implicit intra-class representation direction constraint, resulting in sub-optimal performance. Specifically, to minimize the training loss, the model will constrain the samples with different modality combinations within the same class to generate embeddings with the same direction as the class center vector (Details can be seen in Section~\ref{ana}). This hinders the model from leveraging the specific information of different modality combinations, resulting in insufficient representation learning. As shown in Fig.~\ref{diveristy-compare} (a) and (c), compared to the unimodal model that does not suffer from inter-modal interference, the intra and inter-modality feature diversities of the vanilla subspace-based model decrease significantly. Consequently, the performance of each modality combination will be inferior.

To this end, we propose a general framework Decoupled Multimodal Representation Network (DMRNet) to assist the robust multimodal learning by decoupling the training and inference representations. Specifically, DMRNet estimates a distribution rather than a fixed point in the latent space for each modality combination input. The mean of this distribution serves as the final embedding used for inference, while the embeddings sampled from this estimated distribution are employed for the latter prediction module to calculate the task loss. As a result, the direction constraint passed back from the loss minimization is blocked by the sampled representation, relaxing the constraint on the inference representation. Consequently, the model is allowed to learn non-parallel inference embedding for each modality combination and capture their specific information. As shown in Fig.~\ref{diveristy-compare} (b) and (d), compared with the vanilla subspace-based methods, this helps increase the intra and inter-modality feature diversity significantly. In particular, the variance of the distribution determines the degree of relaxation. A bigger variance will lead to a bigger sample uncertainty to reduce the correlation between the sampled representation and inference representation, contributing bigger representation space.

Besides, subspace-based methods commonly face the challenge of unbalanced training due to the diversity of input modality combinations. To this end, we introduce a hard combination regularizer for DMRNet to handle the unbalanced training. Specifically, the regularizer first mines the hard modality combinations via the variance ranking. Then the regularizer predicts each hard modality combination and calculates their task losses individually. This helps introduce independent gradient paths for hard modality combinations, encouraging the model to pay more attention to them. Different from existing regularization methods~\cite{rfnet,mmformer,mmanet} that use additional modules, the hard combination regularizer shares the parameters with the prediction module of DMRNet, and thus introduces no extra parameters.

Overall, our contributions are summarized as follows,
\begin{itemize}
  \item We reveal the direction constraint of intra-class representation in conventional incomplete multimodal learning methods. It limits the model's representation ability for the specific information of different modality combinations.
  \item We propose a general framework DMRNet for incomplete multimodal learning by decoupling the training and inference representations. This enables the model to learn non-parallel inference embedding for each modality combination, improving the representation ability.
  % \item We propose the hard combination regularizer to mine and regularize the hard modality combination for DMRNet. Compared with existing unimodal regularizers, this provides a more accurate and flexible regularization.
  \item Extensive experiments on multimodal classification and segmentation tasks demonstrate the effectiveness of the proposed DMRNet.
  
\end{itemize}

\section{Related work}

\subsection{Incomplete multimodal learning}

In this work, we focus on the model robustness for incomplete modality data during inference. Specifically, while a variety of sensory modalities can be collected for training, not all of them are always available during testing due to the device~\cite{spcical-gan1,device1} and working condition~\cite{special-hall1,special-hall2}. To this end, many incomplete multimodal learning methods have been proposed and can be roughly categorized into two types: data imputation-based methods and common subspace-based methods.

\textbf{Data imputation-based methods}: These methods first impute the samples or representations of missing modalities and then apply the traditional multimodal algorithm directly. Early works relied on generative adversarial networks to reconstruct the modality sample~\cite{MC2,MC3,spcical-gan1}. Due to the complexity of sample reconstruction, recent methods reduce the problem from input space to latent space to impute the modality representation via generative adversarial networks~\cite{gan-reprentation-1,gan-reprentation-2} or knowledge distillation~\cite{special-hall2,special-hall3,special-hall4,special-hall5,special-hall6}. Although these methods achieve promising results, almost all of them consider the scenarios with two modalities only and are difficult to be scaled to scenarios with multiple modalities. 
 
 % This lim situation are obtained, these methods have to train and deploy a specific model for each missing modality, which has high complexity in practical applications, especially when the modality number is large.

% Generative Multi-View Human Action Recognition

\textbf{Common subspace-based methods}: These methods usually employ the modality dropout technique to produce random input modality combinations and direct the model toward acquiring the invariant feature of all feasible modality combinations. For example, HeMIS~\cite{hemis} generates multimodal embeddings by calculating statistics, including mean and variance, from any quantity of available modalities. Furthermore, Chen~\etal introduce feature disentanglement to eliminate modality-specific information. Recent approaches, such as LCR~\cite{lcr} and RFNet~\cite{rfnet}, concentrate on extracting modality-invariant representation via different attention mechanisms. Additionally, mmFormer~\cite{mmformer} incorporates transformer blocks to model global semantic information for the modality-invariant embedding. Wang~\etal propose the ShaSpec framework to capture the multimodal information with adversarial learning~\cite{shaspec}. While these techniques achieve promising results in complexity and scalability, they contain implicit direction constraints on the intra-class representation vector. This limits the model's ability to capture the specific information of each modality combination, resulting in sub-optimal performance.

% rely solely on deterministic embedding to capture the invariant feature. This 
% and model representation ability

 % so that it can deal with incomplete modality input during inference

% Although these methods achieve promising results in complexity and scalability, all of them equire deterministic embedding to be invariant for complex input variations. This may limit the feature diversities and model representation ability, resulting in sub-optimal performance.

% tend to be limited since the input variation is complex while the embedding needs to be deterministic and invariant. This would lead to a sub-optimal  

% They can be further divided into two types: auto-encoder generation, and attention fusion. Auto-encoder generation guide the model to generate all modality data from any input Correlation maximization

% generative-based method This, however, comes with a large increase in computational cost at test time due to the reconstruction of missing modalities. While this 
% espcifically

% While these latter methods are the state of the art in cross-modal retrieval, they require excessive computation that renders them inapplicable for real- world deployment: for each query, cross-attention is applied to each element in the search database. Our focus is on joint embeddings computed independently in each domain to fa- cilitate large-scale, efficient search.
% Probabilistic

\subsection{Probabilistic Embeddings} Probabilistic representations of data have a long history in machine learning. They were introduced for word embeddings to model the uncertainty about the target concepts with which the input may be associated~\cite{pe-word-1,pe-word-2,pe-word-3}. Recently, probabilistic embeddings have been introduced for vision tasks. Sun~\etal propose view-invariant probabilistic embedding to handle the project uncertainty between 2D and 3D poses~\cite{pe-pose}. Shi~\etal introduce the probabilistic embedding for face recognition system to deal with the quality uncertainty of face image~\cite{pe-face-1}. And Chang~\etal extends it by making the mean of distribution learnable to get better intra-class compactness and inter-class separability~\cite{pe-face-2}. More recent works leverage probabilistic embedding for image-text retrieval to handle the semantic uncertainty of image input~\cite{pe-re-1,pe-re-2}. 

% Generally, these works can be categorized the “one-to-many” tasks. They leverage probabilistic embedding to enhance the representation ability of the model, allowing it to deal with the mapping uncertainty.  In contrast, we utilize probabilistic embedding to introduce the uncertainty to the multimodal representation, decoupling the training and inference representation to relax the direction constraint caused by the intra-class direction constraint and improving the model's representation ability.

% to introduce the uncertainty for the gradient incomplete multimodal learning is a ``many-to-one" task. Given multiple and different modality combination inputs within the same class, it is required to generate the same and accurate output. Here, we introduce the probabilistic embeddings to relieve the representation constraints for richer feature diversities.

% Our method introduces data uncertainty into multi-view learning. With the help of uncertainty, the pro- posed model can automatically estimate the importances of different views for different samples. Superior performance indicates that incorporating data uncertainty in information integration is more suitable to real-world applications.
% Proposed

% \setlength{\belowcaptionskip}{-0.3cm}

\section{Methods}

% In this work, we design DMRNet for incomplete multimodal learning. It consists of three components: decoupled multimodal representation, hard combination regularize, and, target task predictor. Specifically, decoupled multimodal representation help handle the input modality variants for richer feature diversities. In addition, we also propose an hard combination regularizer to further improve the model representation for hard modality combination samples. In this section, we first revisit the formulations of classic deterministic embeddings for incomplete multimodal learning. Then we introduce the details of the proposed DMRNet, including the decoupled multimodal representation and hard combination regularizer. 

\begin{figure*}[t]
\centering
\includegraphics[width=1.0\textwidth]{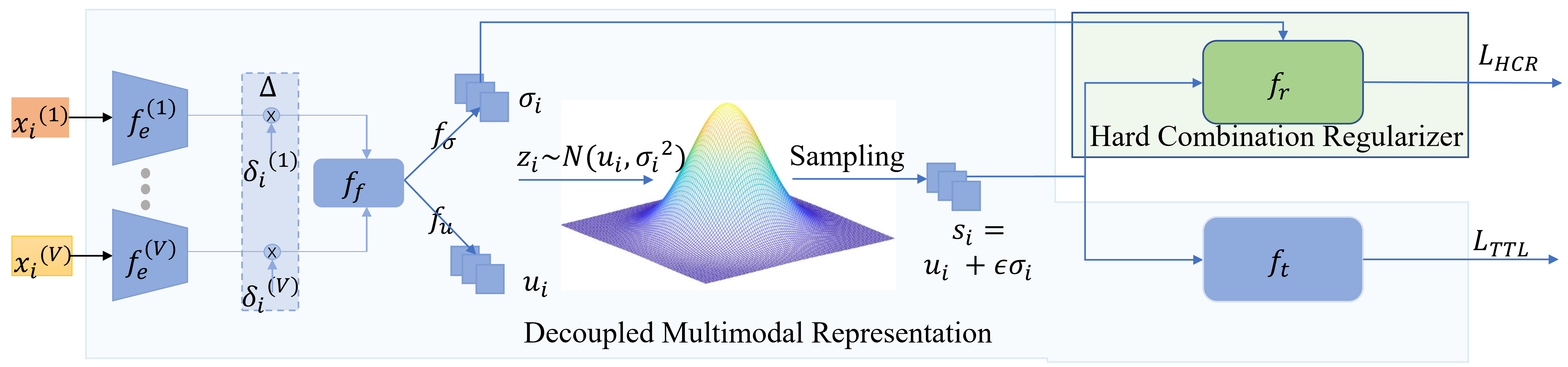} % Reduce the figure size so that it is slightly narrower than the column. Don't use precise values for figure width.This setup will avoid overfull boxes.
\caption{The overall framework of the proposed DMRNet. It consists of two parts: 1) the decoupled multimodal representation that decouples the inference representation and training representation to alleviate the direction constraint on the inference representation; and 2) the hard combination regularizer that mines and regularizes the hard modality combinations to handle the unbalanced training problem.}
\label{DMRNet}
\end{figure*}

In this work, we first analyze the representation constraint of the conventional common subspace-based method. Then we introduce the DMRNet to assist incomplete multimodal learning. As shown in Fig.~\ref{DMRNet}, it consists of two components: decoupled multimodal representation and hard combination regularizer. Specifically, decoupled multimodal representation helps relax the direction constraints on inference representation. The hard combination regularizer further improves the model representation ability for hard modality combination inputs.

 % by decoupling the training and inference embeddings

% This section will introduce the details of them, respectively.

% In this section, we first revisit the formulations of classic deterministic embeddings for incomplete multimodal learning. Then we introduce the details of the proposed DMRNet, including the decoupled multimodal representation and hard combination regularizer. 

\textbf{Notations.} The notations used in this paper are described as follows. $X =\left\{x_{i},y_{i}\right\}_{i=1}^{N}$ is a multimodal dataset that has $N$ samples.  Each $x_i$ consists of $V$ inputs from different modalities as $x_i=({x_i}^{(1)},...,{x_i}^{(V)})$ and $y_i \in[1,2,...,M]$, where $M$ is the number of categories. $\Delta_i=\{{\delta_i}^{(1)},...{\delta_i}^{(V)}\}$ is the Bernoulli sequence corresponding to the $V$ modality encoders, where ${\delta_i}^{(v)} \in \{0,1\}$ is the Bernoulli indicator for the $v_{th}$ encoder. $f_{e}^{(v)}$ denotes the functions for $v_{th}$ modality encoder. $f_{f}$, $f_{\mu}$, and $f_{\sigma}$ denote the functions for calculating the fusing representation $z$, mean value $\mu$, and standard deviation $\sigma$. $f_{t}$, and $f_{r}$ denote the mapping functions for task predictor and hard combination regularizer, respectively. 

% $L_{T}$ is the loss function defined by the target task. 

% $y_{r}$ and $y_{t}$ denote the prediction output of the hard combination regularizer and task predictor, respectively

\subsection{Analysis of Representation Constraint}
\label{ana}

We introduce the analysis of the representational constraint problem for incomplete multimodal learning and find that different input combinations from the same class will be forced to learn the embeddings with the same direction. This limits the model's representation ability for the specific information of different modality combinations, leading to insufficient representation learning. 

Specifically, the multimodal embedding $z_i$ of $x_i$ in the conventional common subspace-based model can expressed as follows,

\begin{numcases}{}
\label{drop}
   z_i=f_{f}(\theta_{f},{r_i}^{(1)}*{\delta_i}^{(1)},...,{r_i}^{(V)}*{\delta_i}^{(V)})\\
  {r_i}^{(v)}=f_{e}^{(v)}({\theta_{e}}^{v},{x_i}^{v}), \quad v\in[1,V]
\end{numcases} where ${r_i}^{(v)}$ is the embedding of the $v_{th}$ modality. $\theta_{f}$ is the parameters for the fusion module. ${\theta_{e}}^{v}$ is the parameters for the $v_{th}$ modality encoder. $ {\delta_i}^{v}  \in [0,1]$ is the  Bernoulli indicator for the $v_{th}$ modality of $x_i$. It is randomly set to either 0 or 1 to simulate random modality missing. This makes the model robust for incomplete inference data. 

 % where $B, C, H, W$ are the batch size, channel number, height, and width, respectively.

Representatively, we take the classification task as an example for analysis here. Let $W \in R^{C\times M}$ denote the parameters of the final linear classifier. The typical cross entropy loss $L_{CE}(.)$ for the  model is defined as follows,

\begin{equation}
    \mathcal{L}_{C E}=-\frac{1}{N} \sum_{i=1}^N \log \frac{e^{W_{y_i}g(z_i)}}{\sum_{k=1}^M e^{(W_{k}g(z_i))}}
\end{equation} where $C$ is the channel dimension $W_{y_{i}}$ means the $y_{x}$-th column of the $W$. $g(z_i)$ denotes the feature vector after the global average pooling and flatten operators.

For a well-trained model, it needs to minimize the loss of each class. Here, $W_{y_{i}}$ is fixed for $y_{i}$-th class. To minimize the loss of class $y_{i}$, $W_{y_i}g(z_i)$ should be maximized. In other words, all input modality combinations belonging to class $y_{i}$ should learn a $g(z_i)$ that has the same direction with $W_{y_i}$.  Consequently, the specificity of each input modality combination will be limited, leading to insufficient representation learning.

\subsection{Decoupled Multimodal Representation}
\label{tp}
As discussed, the representation ability of conventional common subspace-based methods is limited by the intra-class representation constraint for different input combinations. To solve the problem, we introduce the decoupled multimodal representation technique to alleviate the direction constraint on the inference representation by decoupling the training and inference representations. Specifically, it consists of two parts: representation probabilization and distribution regularization.

\textbf{Representation Probabilization}. Different from existing methods that leverage the deterministic embedding in Eq.(~\ref{drop}) for task predictors directly, we consider building probabilistic embeddings, i.e. $z_i\sim p(z_i|x_i)$, for a more flexible representation space. 

For simplicity, we define the probabilistic embedding $z_{i}$ obeys a multivariate Gaussian distribution,

\begin{equation}
\label{gauss}
p(z_i|x_i)=\mathcal{N}(z_i ; \mu_i, {\sigma_i}^{2})
\end{equation} where both the parameters ($\mu_i$ and $\sigma_i$) of the Gaussian distribution are input-dependent predicted. Different from existing methods~\cite{pe-face-1,pe-face-2} that estimate $\mu_{i}$ and $\sigma_{i}$ for the feature vector after pooling, we estimate the $\mu_i$ and $\sigma_i$ for the feature map directly. This not only contributes to better performance but also enables the model to handle the dense prediction task, such as segmentation. For a specific, $\mu_i$ and $\sigma_i$ are defined as follows,

\begin{numcases}{}
  \mu_i=f_{\mu}(\theta_{\mu},z_i) \\
  \log(\sigma_i)=f_{\sigma}(\theta_{\sigma},z_i)
\end{numcases} $\theta_{\mu}$ and $\theta_{\sigma}$ are the parameters for $f_{\mu}(.)$ and $f_{\sigma}(.)$, respectively. Here we predict $\log\sigma_i$ instead of $\sigma_i$ directly for a better stability~\cite{pe-re-1,pe-re-2}. In detail, we implement $f_{\mu}(.)$ and $f_{\sigma}(.)$ with a simple 1x1 convolution with the batch norm, respectively, which introduces negligible parameters.

% is not a deterministic embedding anymore, but

Now, the representation of each sample becomes a stochastic embedding sampled from $\mathcal{N}(z_i; \mu_i, {\sigma_i}^{2})$. Nevertheless, the sampling operation is not differentiable. Thus, we consider the reparameterization trick~\cite{rep} to enable backpropagation,

% consider the embedding as a feature map and
% treat the probabilistic embedding as a vector and estimate the $\mu$ and $\sigma$

% we consider the embedding as a feature map

% ethods usually the probabilistic embedding is treated as a vector and , $f_{\mu}$ and $f_{\sigma}$ usually We use a simple 1x1 convolution with batch norm to implement $f_{\mu}$ and $f_{\sigma}$, respectively. Compared with existing methods~\cite{pe-re-1,pe-re-2} that treat the probabilistic embedding as a vector and estimate the $\mu$ and $\sigma$ with a fully-connect layer, This not only reduces the computational complexity but also allows the decoupled multimodal representation to handle the dense prediction task, such as segmentation.

\begin{equation}
\label{e-rep}
{s_i}={\mu_i}+\epsilon {\sigma_i}, \quad \epsilon \sim \mathcal{N}(\mathbf{0}, \mathbf{I})
\end{equation}

In general, we sample noise from $\mathcal{N}(\mathbf{0}, \mathbf{I})$ and obtain the embedding $s_i$ following Eq.(\ref{e-rep}) instead of directly sampling from $\mathcal{N}(z_i; \mu_i, {\sigma_i}^{2})$. Here, $s_i$ is the embedding for the training of the prediction module and $\mu_{i}$ is the final embedding for inference. 

% $\sigma$ controls the sampling range and can be regarded as the relaxation coefficient.

In this way, DMRNet decouples the training and inference embedding. And the cross-entropy loss for the model can be rewritten as follows,

\begin{equation}
    \mathcal{L}_{C E}=-\frac{1}{N} \sum_{i=1}^N \log \frac{e^{W_{y_i}g(s_i)}}{\sum_{k=1}^M e^{(W_{k}g(s_i))}}
\end{equation}

This only requires the sampled embedding $g(s_{i})$ to share the same direction with $W_{y_i}$. The inference embedding $\mu_{i}$ of different input combinations belonging to the same class could be non-parallel. This relaxes the direction constraint on the inference representation and enables the model to capture the specific information for different modality combinations. In particular, the value of the $\sigma_{i}$ controls the degree of relaxation. When  $\sigma_{i}=0$, $s_i$ will equal $\mu_i$, which degenerates into the vanilla subspace-based methods without relaxation. In contrast, bigger $\sigma_{i}$ means a bigger sampling range and the direction constraint from $s_i$ passed to $\mu_i$ will become weaker.

\textbf{Distributional Regularization} As discussed, bigger $\sigma_{i}$ will lead to weaker constraints. Therefore, the model tends to predict a bigger $\sigma_{i}$ to improve the model's ability to capture the specific information of different modality combinations, improving the representation ability. However, bigger $\sigma_i$ will also introduce high uncertainty, which would hinder the optimization of $W$. Therefore, we need to introduce a regularization term for $\sigma_i$ to limit its range. Inspired by previous probabilistic embedding methods~\cite{pe-face-1,pe-face-2}, we introduce a regularization term during the optimization to explicitly constrain the distance between $\mathcal{N}(z; \mu_i, {\sigma_i}^{2})$ and normal Gaussian distribution, $\mathcal{N}(\mathbf{0}, \mathbf{I})$,

\begin{equation}
\begin{aligned}
L_{DR} & =\frac{1}{N}\sum_{i=1}^{N}K L\left[\mathcal{N}\left(z_i \mid {\mu_i}, {\sigma_i}^2\right) \| \mathcal{N}(\epsilon \mid \mathbf{0}, \mathbf{I})\right] \\
& =-\frac{1}{N}\sum_{i=1}^{N}\frac{1}{2}\left(1+\log {\sigma_i}^2-{\mu_i}^2-{\sigma_i}^2\right)
\end{aligned}
\end{equation} where $KL(.||.)$ means the KL divergence.

 % to control the effect of distributional regularization for target task learning. A detailed analysis of them can be seen in the experiments.

% Given the sample representation $s$ from input $x$,

% \textbf{Comparison with existing probabilistic embedding methods}. There are two key differences between the decoupled multimodal representation and existing probabilistic methods~\cite{pe-re-1,pe-re-2,pe-face-1}. 1)Existing methods tend to estimate the $\mu$ and $\sigma$ for each modality. In contrast, we estimate $\mu$ and $\sigma$ from the fused feature directly. This allows the proposed decoupled multimodal representation to be easily integrated into existing multimodal models. 2) Existing methods mainly consider the classification task and estimate the $\mu$ and $\sigma$ for the pooled feature vector with fully-connect layer. Here, we consider estimating the $\mu$ and $\sigma$ for the feature map before pooling with the convolutional neural network. This not only allows the decoupled multimodal representation to handle the dense prediction task, such as segmentation, but improves the model performance.

\subsection{Hard Combination Regularizer}
\label{udr}
Since DNNs tend to first memorize simple samples before overfitting hard samples~\cite{mem1,mem2,mem3}. The common subspace model tends to fit the discriminative modality combinations first. Consequently, their performance will significantly decline when facing hard modality combinations. To handle this unbalanced training issue, we introduce the hard combination regularizer. It mines the hard modality combination relying on the estimated $\sigma_i$  and then introduces independent gradient paths for them to regularize their optimization. Compared with the traditional regularization methods~\cite{rfnet,mmformer,mmanet} that use additional modules, this regularizer shares the parameters with the prediction module of DMRNet, and thus introduces no extra parameters.

 % the optimization of hard combinations will be insufficient during the early training stage. Thus,
 % that of

\textbf{Hard Combinations Mining}. Since the model tends to fit the input from discriminative modality combinations first, the hard combinations will have a bigger variance than the discriminative combinations that are well optimized by $L_{DR}$. And we can mine the hard combinations through simple variance ranking.

Specifically, given $V$ modalities, we can obtain $2^V$ modality combinations by setting the components in $\{{\delta_i}^{(1)},...{\delta_i}^{(V)}\}$ as 0 or 1 randomly. Here, we calculate the variance set $\mathcal{D}=\{d_1,...,d_{2^{V-1}}\}$ of all combinations among the training dataset,

% \begin{equation}
%   \label{variance}
%     d_j=\frac{1}{N_j*c*h*w}\sum\limits_{N_j}\sum\limits_{c}\sum\limits_{h}\sum\limits_{w}(f_{\sigma}^{2}(f_{f}({r_i}^{(1)}*\delta^{(1)},...,{r_i}^{(V)}*\delta^{(V)})))
% \end{equation}

\begin{equation}
  \label{variance}
    d_j=\frac{1}{N_j*c*h*w}\sum_{p=1}^{N_j}\sum\limits_{c}\sum\limits_{h}\sum\limits_{w}({\sigma_{p}}^2)
\end{equation} where $j=\sum (2^{k-1})*\delta^{k}, 1<=k<=V$, $N_j$ denotes the number of the $j_{th}$ combination, $c$, $h$, $w$ denote the channel, height and width of the estimated variance.

For a fair comparison with existing methods, we also select $V$ combinations as the hard ones for regularization. Thus the hard combinations set $\mathbf{H}$ consisting of the combinations $\Delta_j$ that have the top-V variance in $\mathcal{D}$.

\textbf{Hard Combinations Regularization} As shown in Fig.~\ref{DMRNet}, to encourage the model to pay more attention to the hard modality combination input, we introduce an auxiliary predictor to guide it to decide the hard modality combination inputs independently. 

Specifically, given a sampling embedding $s_i$ coming from the modality combination $\Delta_{i}$, it calculates the hard combination regularizer loss $L_{HCR}$ for $x_{i}$ as follows, 

% \begin{equation}
%   L_{UDR}=L_{T}(f_{r}(s))
% \end{equation}

% $y_{s}=f_{r}(s)$ if $\Delta$ ,

% \begin{numcases}{y_r=}
%   f_{r}(s) \quad \ \Delta \in \mathbf{H}\\
%   \quad0 \quad\quad \Delta \not\in \mathbf{H}
% % \end{numcases} 

\begin{numcases}{L_{HCR}=}
  L_{CE}(f_{r}(g(s_i)),y_i) \quad \ \Delta_i \in \mathbf{H}\\
  \quad0 \quad\quad\quad\quad\quad \Delta_I \not\in \mathbf{H}
\end{numcases} where $f_{r}$ is the mapping function that converts the embedding into output required by the task loss function. It shares the parameter with the $f_{t}$ in the task predictor.

\subsection{Total Loss}
The total training loss $L$ for the DMRNet is defined as follows,
\begin{equation}
\begin{aligned}
  L = L_{TTL}+\alpha L_{DR} + \beta L_{HCR}. 
\end{aligned}
\end{equation} where $L_{TTL}$ is the traditional target task loss for input data. $\alpha$ and $\beta$ are the hyper-parameters. Specifically, $\alpha$ controls the level of relaxation, and $\beta$ controls the regularization degree for hard modality combinations. A detailed analysis of them can be seen in Section~\ref{pa}.

\subsection{Relationship to Prior Work} 
Some concurrent works~\cite{pe-face-1,pe-re-1} seem similar to our DMRNet, which also utilize the technique of probabilistic representation. However, there are still differences in motivations and implementations with ours. They leverage probabilistic embedding to deal with the mapping uncertainty caused by input variants. In contrast, we utilize probabilistic embedding to introduce the uncertainty to the multimodal representation proactively, decoupling the training and inference representation to relax the direction constraint caused by the intra-class direction constraint. 

Besides, they estimate the mean and variance for the 1D feature vectors, however, we estimate them for the 2D feature maps directly. This not only contributes to better performance (see ablation experiments in Table~\ref{ab-pe}) but also enables DMRNet to handle the dense prediction task (see ablation experiments in Table~\ref{pe-se-nyu}).

\section{Experiments}

\begin{table*}[h]

\caption{Performance on the multimodal classification task with CASIA-SURF. The metric is ACER($\downarrow$) and the lower the value, the better the performance. $\CIRCLE$ and $\Circle$ denote available and missing modalities, respectively.}
\label{pe-c-surf}

\centering

\begin{tabular}{ccccccccccc}
\toprule
\multicolumn{3}{c|}{Modality}                & \multicolumn{7}{c}{Methods}                                                                     \\ \toprule
\multicolumn{1}{c}{RGB} & \multicolumn{1}{c}{Depth} & IR & \multicolumn{1}{|c}{SF-MD}  & \multicolumn{1}{c}{HeMIS} & \multicolumn{1}{c}{LCR}  & \multicolumn{1}{c}{RFNet} & \multicolumn{1}{c}{mmFormer} & \multicolumn{1}{c}{ShaSpec}& MMANet& DMRNet    \\ \toprule
\multicolumn{1}{c}{\CIRCLE} & \multicolumn{1}{c}{\Circle}   &\Circle& \multicolumn{1}{|c}{11.75}  & \multicolumn{1}{c}{14.36} & \multicolumn{1}{c}{13.44} & \multicolumn{1}{c}{12.43} & \multicolumn{1}{c}{11.15} & \multicolumn{1}{c}{11.57} & 8.57& \textbf{8.23} \\ 
\multicolumn{1}{c}{\Circle}  & \multicolumn{1}{c}{\CIRCLE}  &\Circle& \multicolumn{1}{|c}{5.87}  & \multicolumn{1}{c}{4.70} & \multicolumn{1}{c}{4. 40} & \multicolumn{1}{c}{4.17}  & \multicolumn{1}{c}{3.67} & \multicolumn{1}{c}{6.25}  & 2.27& \textbf{2.01} \\ 
\multicolumn{1}{c}{\Circle}  & \multicolumn{1}{c}{\Circle}   & \CIRCLE & \multicolumn{1}{|c}{16.62} & \multicolumn{1}{c}{16.21} & \multicolumn{1}{c}{15.26} & \multicolumn{1}{c}{14. 69} & \multicolumn{1}{c}{13.99} & \multicolumn{1}{c}{10.71}  & 10.04& \textbf{8.98} \\ 
\multicolumn{1}{c}{\CIRCLE} & \multicolumn{1}{c}{\CIRCLE}  &\Circle& \multicolumn{1}{|c}{4.61}  & \multicolumn{1}{c}{3.23} & \multicolumn{1}{c}{3.32} & \multicolumn{1}{c}{2.23}  & \multicolumn{1}{c}{1.93} & \multicolumn{1}{c}{3.11}  & 1.61& \textbf{1.21} \\ 
\multicolumn{1}{c}{\CIRCLE} & \multicolumn{1}{c}{\Circle}   & \CIRCLE & \multicolumn{1}{|c}{6.68}  & \multicolumn{1}{c}{6.27} & \multicolumn{1}{c}{5.16} & \multicolumn{1}{c}{4.27}  & \multicolumn{1}{c}{4.77} & \multicolumn{1}{c}{4.23}  & 3.01& \textbf{3.00} \\ 
\multicolumn{1}{c}{\Circle}  & \multicolumn{1}{c}{\CIRCLE}  & \CIRCLE & \multicolumn{1}{|c}{4.95}  & \multicolumn{1}{c}{3.68} & \multicolumn{1}{c}{3.53} & \multicolumn{1}{c}{3.22}  & \multicolumn{1}{c}{3.10}  & \multicolumn{1}{c}{2.52} & 1.18& \textbf{0.80} \\ 
\multicolumn{1}{c}{\CIRCLE} & \multicolumn{1}{c}{\CIRCLE}  & \CIRCLE & \multicolumn{1}{|c}{2.21}  & \multicolumn{1}{c}{1.97} & \multicolumn{1}{c}{1.88} & \multicolumn{1}{c}{1.18}  & \multicolumn{1}{c}{1.94}  & \multicolumn{1}{c}{1.79} & 0.87 & \textbf{0.66} \\ \toprule
\multicolumn{3}{c}{Average}                & \multicolumn{1}{|c}{7.52}  & \multicolumn{1}{c}{7.18} & \multicolumn{1}{c}{6.71} & \multicolumn{1}{c}{6.02}  & \multicolumn{1}{c}{5.93} & \multicolumn{1}{c}{5.59} &3.94 & \textbf{3.58} \\ \toprule
\end{tabular}
\end{table*}
\vspace{-0.0em}

\subsection{Performance and Comparison on Face Anti-spoofing Task}

\textbf{Datasets:} We performed experiments on CASIA-SURF~\cite{surf} dataset. It contains the RGB, Depth, and IR modalities. To evaluate the performance of DMRNet, we followed the intra-testing protocol suggested by the authors for CASIA-SURF, dividing it into train, validation, and test sets with 29k, 1k, and 57k samples, respectively. 

% and CeFA~\cite{cefa}

% Similarly, for CeFA, we followed the cross-ethnicity and attack protocol recommended by the authors and divided it into train, validation, and test sets with 35k, 18k, and 54k samples, respectively. The reported results are based on the test set and measured using the Average Classification Error Rate (ACER)~\cite{surf}.

\textbf{Comparison:} Here we compare DMRNet with two groups of common subspace-based methods. One focuses on extracting modality-invariant features, such as HeMIS~\cite{hemis} and LCR~\cite{lcr}. Another considers improving the discrimination ability for hard combinations, such as mmFormer~\cite{mmformer}. Besides, we also consider the recent SOTA methods, ShaSpec~\cite{shaspec} and MMANet~\cite{mmanet}.

To establish a baseline, we introduced the SF-MD method. It is a variant of the benchmark method on CASIA-SURF~\cite{surf} by adding the Bernoulli indicator $\delta$ after each modality encoder. The metric is the Average Classification Error Rate (ACER)~\cite{surf}.

% For a fair comparison, we unify the implementation of all comparison methods as SF-MD.

\textbf{Implementation}: We unify the backbone and fusion module as ResNet18 and the convolution block followed by the benchmark~\cite{surf} for a fair comparison. We use random flipping, rotation, and cropping for data augmentation. All models are optimized by an SGD for 100 epochs with a mini-batch 64. The learning rate was initialized to 0.001 with a linear warm-up of five epochs and was divided by 16 at 33 and 50 epochs. Weight decay and momentum are set to 0.0005 and 0.9, respectively.

% To establish a baseline, we introduced the SF-MD method, a variant of the benchmark method on CASIA-SURF~\cite{surf}. SF-MD adds the Bernoulli indicator $\delta$ after each modality encoder of the benchmark method to accommodate incomplete multimodal input modality combinations.

% For a fair comparison, we unify the implementation of all comparison methods as that of the benchmark method of CASIA-SURF. The details can be seen in the supplementary material.

\textbf{Results:} Table~\ref{pe-c-surf} shows the comparison results with the state-of-the-art methods on the CASIS-SURF dataset. We can see that DMRNet achieves the best performance for all the nine modality combinations consistently. This demonstrates the effectiveness of the proposed DMRNet for the incomplete multimodal classification task. Particularly, while DMRNet does not show significant performance gain to MMANet, it and other competitors require no assistance and thus can handle the modality missing appearing in the training phase. In contrast, MMANet requires complete multimodal data to train the teacher model, limiting its application. More specifically, under the setting without model assistance, DMRNet shows significant performance improvements compared to the SOTA method, ShaSpec, by 2.01\%.

% Particularly, while MMANet has a similar performance to DMRNet, it requires complete multimodal data to train the teacher model. In contrast, DMRNet and other competitors require no assistance and thus can handle the modality missing appearing in the training phase. In other words, under the setting without model assistance, DMRNet shows significant performance improvements compared to the SOTA method, ShaSpec, by 2.01\%.

% Particularly, DMRNet outperforms the recent SOTA methods, ShaSpec and MMANet, by 2.01\% and 0.36\% on average, respectively.

% Here, while DMRNet does not show significant performance gain to MMANet, DMRNet has a similar performance to DMRNet, it requires complete multimodal data to train the teacher model. In contrast, DMRNet and other competitors require no assistance and thus have better applicability, for example, when a certain modality of data is missing during the training phase.

  % Compared with the second-best method, i.e. ShaSpec, DMRNet decreases the average ACER by 1.30\% on the CASIS-SURF dataset. Besides, we can see that DMRNet achieves the best performance for all the nine modality combinations.

% This shows the superiority of our method on the incomplete multimodal classification task. More importantly, DMRNet even outperforms the customized baseline method, i.e. SF, for all the modality combinations on the CASIA-SURF and CeFA, decreasing the average ACER by 1.86\% and 1.66\%. 

\begin{table}[]
\centering
\caption{Performance on the multimodal emotion recognition task with the CREMA-D and Kinetics-Sounds datasets. The metric is accuracy($\uparrow$) and the higher the value, the better the performance. $\CIRCLE$ and $\Circle$ denote available and missing modalities, respectively.}
\label{emo-com}

\setlength{\tabcolsep}{1.5mm}{
\begin{tabular}{cccccc}
\toprule
\multicolumn{2}{c|}{Modalities}                          & \multicolumn{3}{c}{Methods}                                                        \\ \hline
\multicolumn{1}{c}{Audio} & \multicolumn{1}{c|}{Visual} & \multicolumn{1}{c}{OGM-MD} & \multicolumn{1}{c}{ShaSpec}  &MMANet      & DMRNet         \\ \hline\hline
\multicolumn{6}{c}{CREMA-D}                                                                                                                   \\ \hline
\multicolumn{1}{c}{\CIRCLE}   & \multicolumn{1}{c|}{\Circle}       & \multicolumn{1}{c}{59.13}  & \multicolumn{1}{c}{\textbf{59.86}}     &58.89     & 58.87 \\ 
\multicolumn{1}{c}{\Circle}      & \multicolumn{1}{c|}{\CIRCLE}    & \multicolumn{1}{c}{41.66}  & 47.17 & 47.31 & \textbf{55.10}          \\ 
\multicolumn{1}{c}{\CIRCLE}   & \multicolumn{1}{c|}{\CIRCLE}    & \multicolumn{1}{c}{63.03}  & \multicolumn{1}{c}{66.80}  & 67.87       & \textbf{70.10} \\ \hline
\multicolumn{2}{c|}{Average}                             & \multicolumn{1}{c}{55.69}  & \multicolumn{1}{c}{57.94}     &   57.66  & \textbf{61.35} \\ \hline\hline
\multicolumn{6}{c}{Kinect-Sound}                                                                                                             \\ \hline
\multicolumn{1}{c}{\CIRCLE}   & \multicolumn{1}{c|}{\Circle}       & \multicolumn{1}{c}{48.00}  & \multicolumn{1}{c}{45.98}    &    49.11   & \textbf{51.56} \\
\multicolumn{1}{c}{\Circle}      & \multicolumn{1}{c|}{\CIRCLE}    & \multicolumn{1}{c}{40.51}   & \multicolumn{1}{c}{45.43}     &42.06     & \textbf{48.13} \\ 
\multicolumn{1}{c}{\CIRCLE}   & \multicolumn{1}{c|}{\CIRCLE}    & \multicolumn{1}{c}{63.74}  & \multicolumn{1}{c}{63.45}     & 64.66     & \textbf{68.34} \\ \hline
\multicolumn{2}{c|}{Average}                             & \multicolumn{1}{c}{50.74}  & \multicolumn{1}{c}{51.62}     &51.94     & \textbf{56.01}          \\ \toprule
\end{tabular}
}
\end{table}
% \vspace{-3.0em}

\subsection{Performance and Comparison on Audio-Visual Recognition Task} 

\textbf{Datasets}: We perform experiments on CREMA-D and Kinetics-Sounds datasets. Both of them contain audio and visual modalities. This helps to evaluate the generalization of the proposed DMRNet to non-visual modalities. Specifically, CREMA-D contains 7,442 video clips for 6 common emotions. Following its benchmark, the whole dataset is randomly divided into a 6698-sample training set and a 744-sample testing set. Besides, Kinetics-Sounds is a large-scale dataset that contains 19k 10-second video clips for 34 actions (15k training, 1.9k validation, 1.9k test).

\textbf{Comparison}: Here, we set OGM-MD as the baseline, which is the variant of OGM~\cite{ogm} by adding the Bernoulli indicator $\delta$ after the modality encoder. We mainly compare DMRNet with the recent SOTA method, ShaSpec~\cite{shaspec} and MMANet~\cite{mmanet}. The metric is classification accuracy.

% For a fair comparison, we unify the implementation of all comparison methods as that of OGM. 

\textbf{Implementation} We unify the backbone and fusion module as those of OGM~\cite{ogm} for a fair comparison. Specifically, for the visual encoder, we take 1 frame as input; for the audio encoder, we slightly change the input channel of ResNet18 from 3 to 1, and the rest parts remain unchanged. We process the audio dataset into a spectrogram of size 257×299 for CREMA-D and 257×1,004 for Kinetics-Sounds. We use SGD with 0.9 momentum and 1e-4 weight decay as the optimizer. The learning rate is 1e-3 initially and multiplies 0.1 every 70 epochs.

\textbf{Result}: As shown in Table~\ref{emo-com}, DMRNet also achieves the best average performance on both datasets. Specifically, it outperforms the ShaSpec and MMANet by 3.41\% and 3.69\%, respectively, in CREMA-D and by 4.39\% and 4.07\%, respectively in Kinetics-Sounds.  This verifies its effectiveness on non-visual modality tasks. Moreover, compared to the CREMA-D, the performance gain of DMRNet on Kinetics-Sounds is higher, showing its superiority in processing large-scale datasets.

\subsection{Ablation Study}
\label{abs}

% In this section, we will study the effectiveness of the decoupled multimodal representation and hard combination regularizer. Then we further explore the effectiveness of distributional representation in the decoupled multimodal representation. we conduct extensive ablation experiments on all used datasets. Limited by page, we only present the results of the CASIA-SURF dataset and other results can be seen in the supplementary material. 

In this section, we first detail the definition of feature diversity. Then we study the effectiveness of the decoupled multimodal representation and hard combination regularizer. Besides, we compare different implementations of distribution estimation and study the generalization of DMRNet to dense prediction tasks.

% three datasets. Due to space limitations, we solely present the outcomes of the CASIA-SURF dataset, and other findings are available in the supplementary material.

\textbf{Metric for feature diversity}. Since the feature diversity is the channel diversity of the feature maps~\cite{channel}, we introduce the inter-channel distance metrics $D_{channel}$ to evaluate the intra-modality and inter-modality feature diversity. Specifically, given the modality embedding $F_{v}=f_{e}^{v}(x^{(v)}),v\in [1,V]$, we reshape its shape as $R^{c\times h*w}$, where $c, h, w$ are the channel number, height, and width of the embedding. The metric is defined as follows,
\begin{numcases}{}
D_{channel}{[k,:]}=1- ({F_{o}}_{[k,:]}/ \left\|{F_{o}}_{[k,:]}\right\|_{2})\\
F_{o}= F_{m}*(F_{n})^\mathsf{T}, m,n\in [1,V]  
\end{numcases} where the notation [k, :] denotes the $k_{th}$ row in a matrix. If $m=n$, we can get the intra-modality feature diversity that calculates the cosine distance between the channel coming from $n_{th}$ modality. if $m\not=n$, we can get the inter-modality feature diversity that calculates the cosine distance between the channel coming from $m_{th}$ and $n_{th}$ modalities. Generally, higher $D_{channel}$ means higher channel difference, meaning richer feature diversity.

\begin{table}[]
\caption{Ablation results on the CASIA-SURF dataset. `SF-MD' is the baseline model. `DMR' means the decoupled multimodal representation. `HCR' means the hard combination regularizer.}
\label{ab-mad-surf}
\centering
\setlength{\tabcolsep}{2mm}{
\begin{tabular}{ccc|ccc}
\toprule
\multicolumn{1}{c}{RGB} & \multicolumn{1}{c}{Depth} & IR & \begin{tabular}[c]{@{}c@{}}SF-MD\end{tabular} & \begin{tabular}[c]{@{}c@{}}+DMR\end{tabular} & \begin{tabular}[c]{@{}c@{}}+DMR+HCR\end{tabular} \\ \toprule
\multicolumn{1}{c}{\CIRCLE} & \multicolumn{1}{c}{\Circle}   &\Circle& 11.75                        & 9.86(+1.59)                      & \textbf{8.23(+1.63)}                     \\ 
\multicolumn{1}{c}{\Circle}  & \multicolumn{1}{c}{\CIRCLE}  &\Circle& 5.87                         & 2.75(+3.12)                    & \textbf{2.01(+0.74)}                     \\ 
\multicolumn{1}{c}{\Circle}  & \multicolumn{1}{c}{\Circle}   & \CIRCLE & 16.62                       & 10.60 (+6.02)                      & \textbf{8.98(+1.62)}                         \\ 
\multicolumn{1}{c}{\CIRCLE} & \multicolumn{1}{c}{\CIRCLE}  &\Circle& 4.61                         & 2.18(+1.24)                     & \textbf{1.21(+0.97)}                        \\ 
\multicolumn{1}{c}{\CIRCLE} & \multicolumn{1}{c}{\Circle}   & \CIRCLE & 6.68                         & 4.00(+2.68)                     & \textbf{3.00(+1.00)}                     \\ 
\multicolumn{1}{c}{\Circle}  & \multicolumn{1}{c}{\CIRCLE}  & \CIRCLE & 4.95                         &1.61(+3.34)                      &  \textbf{0.80(+0.81)}                     \\ 
\multicolumn{1}{c}{\CIRCLE} & \multicolumn{1}{c}{\CIRCLE}  & \CIRCLE & 2.21                         & 1.43(+0.78)                    & \textbf{0.66(+0.77)}                          \\ \toprule
\multicolumn{3}{c|}{Average}                & 7.52                         & 4.63(+2.89)                       & \textbf{3.58(+0.95)}                     \\ \toprule
\end{tabular}}
\end{table}

% \subsubsection{The effect of decoupled multimodal representation} 

\textbf{The effect of decoupled multimodal representation}. We conduct comparison experiments with the vanilla SF-MD and its variant with decoupled multimodal representation (DMR). The results are shown in Table~\ref{ab-mad-surf}. We can see that DMR improves the performance of vanilla SF-MD by 2.89\% on average. More importantly, DMR brings performance gain on all modality combinations consistently. This demonstrates the effectiveness of decoupling the training and inference representation to alleviate the representation constraints, which enables the model to capture modality-specific information for each modality combination. 

% More importantly, the distribution of feature maps allows PME to handle the dense prediction task, such as segmentation (see Section~\ref{segmemtation}). 

% Furthermore, the proposed PME outperforms PE by 0.94\%, which demonstrates the validity of learning feature map distribution. This is because it considers more embedding parameters

% We study the effectiveness of decoupled multimodal representation with quantitative and visual comparison. w
% \textbf{Quantitative Comparison}. 

% \textbf{Visual Comparison}. 

% We further count and visualize the intra-modality and inter-modality channel differences. As shown in Fig.~\ref{inter}, compared with deterministic point embedding, decoupled multimodal representation increases not only the intra-modality channel differences for RGB (a), Depth (b), and IR modalities (c), but also the inter-modality channel differences of RGB and Depth modalities (d), RGB and IR modalities (e), as well as Depth and IR modalities (f). This confirms the effect of PME to increase feature diversity.

We further visualize the intra-modality and inter-modality channel differences. As examples in Fig.~\ref{diveristy-compare} (b) and (d), compared with the vanilla SF-MD model, DMR increases not only the intra-modality channel diversity of RGB encoder but also the inter-modality channel diversity between RGB and Depth encoder. This further demonstrates that DMR can improve the representation ability for modality-specific information to get richer feature diversity.

\textbf{The effect of hard combination regularizer}. To study the effect of hard combination regularizer, we conduct experiments to compare the performance of the SF-DMR, namely the SF-MD with the decoupled multimodal representation, and its variant with hard combination regularize(HCR). As shown in Table~\ref{ab-mad-surf}, HCR brings improvement on each modality combination consistently and improves the performance of SF-DMR by 0.95\% on average. This demonstrates the effectiveness to regularize the hard input modality combinations. 

% Here the average gain of SR and MAR is less than SP and MAD since they aim to improve the performance of only the weak, not all combinations.

\begin{SCfigure}[0.40][t]
\centering
\includegraphics[width=0.59\columnwidth]{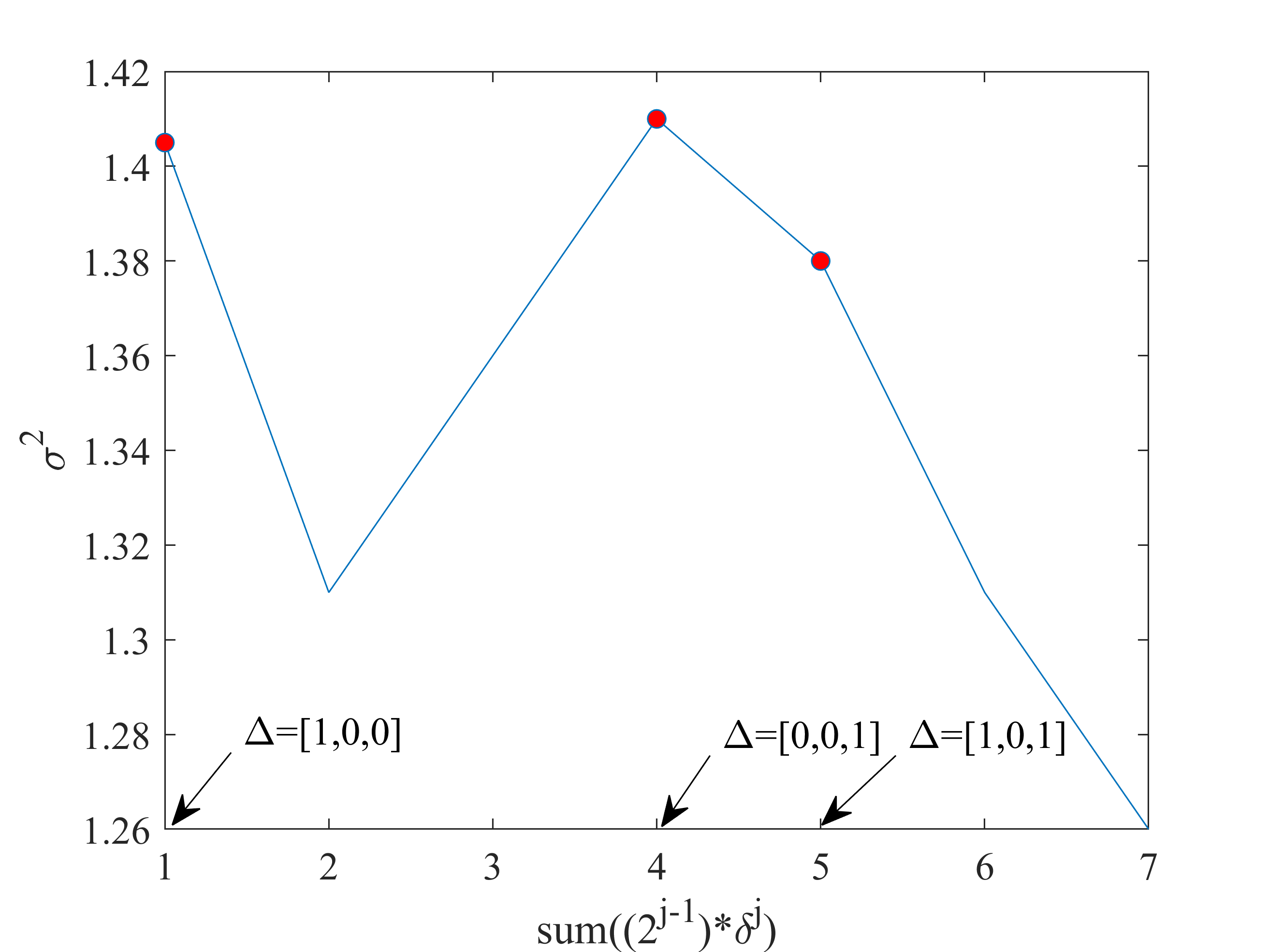} % Reduce the figure size so that it is slightly narrower than the column. Don't use precise values for figure width.This setup will avoid overfull boxes.
\caption{The training variance $\sigma^{2}$ of different modality combinations for CASIA-SURF dataset. The top $V(V=3)$ variances are marked with red circles. $\Delta=\{{\delta}^{(1)},...{\delta_i}^{(V)}\}$ is the Bernoulli sequence corresponding to the $V$ modality encoders.}
\label{training_pro}
% \vspace{-0.5em}
\end{SCfigure}

Specifically, as shown in Fig.~\ref{training_pro}, the largest three training variances come from combinations 1 ($\Delta=[1,0,0]$), 4($\Delta=[0,0,1]$), 5($\Delta=[1,0,1]$), which are exactly the combination of RGB, IR, and RGB+IR. They are also the three worst-performing combinations of SF-DMR (see Table~\ref{ab-mad-surf}). Moreover, the performance gain of HCR mainly comes from RGB (1.63\%), IR (1.62\%), as well as the combination of RGB and IR (1.00\%). These results show that HCR can mine the hard modality combinations accurately and force the DMRNet to improve its discrimination ability for them.

% Specifically, as shown in Table~\ref{ab-mad-surf}, the three worst-performing combinations of SF-PME are `RGB', `IR' and, `RGB+IR'. However, UDR only focuses on the combinations of a single modality and improves the performance of combinations of RGB, Depth, and IR by 0.33\%, 0.75\%, and 0.31\%, respectively. Here `Depth' is exactly a simple modality combination. In contrast, UDR can mine the accurate hard input modality combinations according to the training variance in Eq.~\ref{variance}. As shown in Fig.~\ref{training_pro}, the largest three training variances come from combinations 1 ($\Delta=[1,0,0]$), 4($\Delta=[0,0,1]$), 5($\Delta=[1,0,1]$), which are exactly the combination of RGB, IR, and RGB+IR. And the performance gain of UDR mainly comes from RGB (1.13\%), IR (2.22\%), as well as the combination of RGB and IR (1.17\%). These results show that UDR can mine the hard modality combinations more accurately and force the DMRNet to improve its discrimination ability for them.

% \begin{figure}
% \centering
% \includegraphics[width=1.0\columnwidth]{picture/intra-rgb-3.png} % Reduce the figure size so that it is slightly narrower than the column. Don't use precise values for figure width.This setup will avoid overfull boxes.
% \caption{Comparison of intra-modal (a,b,c) and inter-modal (d,e,f) channel differences in the CASIA-SURF dataset. `Point' denotes the results of the deterministic method, SF-MD. `PME' denotes the results of the decoupled multimodal representation method, DMRNet. For a fair comparison, here DMRNet does not use the hard combination regularizer. ``A:B'' denotes the comparison of average $D_{channel}$ of `Point' and `PME' results.}
% \label{inter}
% % \vspace{-1.0em}
% \end{figure}

% \subsection{Parameters Analysis}

\begin{table}[]
\caption{Performance comparison of different methods to estimate the mean and variance of the probabilistic distribution.}
\label{ab-pe}
\centering
\setlength{\tabcolsep}{2.4mm}{
\begin{tabular}{ccc|cccc}
\toprule
\multicolumn{1}{c}{RGB} & \multicolumn{1}{c}{Depth} & IR & \begin{tabular}[c]{@{}c@{}}SF-MD\end{tabular} & \begin{tabular}[c]{@{}c@{}}+PE\end{tabular} & \begin{tabular}[c]{@{}c@{}}+PCME\end{tabular} & +DMR \\ \toprule
\multicolumn{1}{c}{\CIRCLE} & \multicolumn{1}{c}{\Circle}   &\Circle& 11.75                        & 12.30                        & 10.49   & \textbf{9.86}                  \\ 
\multicolumn{1}{c}{\Circle}  & \multicolumn{1}{c}{\CIRCLE}  &\Circle& 5.87                         & 4.07                       & 3.42        &\textbf{2.75}             \\ 
\multicolumn{1}{c}{\Circle}  & \multicolumn{1}{c}{\Circle}   & \CIRCLE & 16.62                       & 14.31                        & 12.93     &    \textbf{10.60}                \\ 
\multicolumn{1}{c}{\CIRCLE} & \multicolumn{1}{c}{\CIRCLE}  &\Circle& 4.61                         & 3.54                       & 2.98  & \textbf{2.18}                       \\ 
\multicolumn{1}{c}{\CIRCLE} & \multicolumn{1}{c}{\Circle}   & \CIRCLE & 6.68                         & 6.39                       & 4.57  & \textbf{4.00}                     \\ 
\multicolumn{1}{c}{\Circle}  & \multicolumn{1}{c}{\CIRCLE}  & \CIRCLE & 4.95                         &3.81                       & 3.61          & \textbf{1.61}           \\ 
\multicolumn{1}{c}{\CIRCLE} & \multicolumn{1}{c}{\CIRCLE}  & \CIRCLE & 2.21                         & 3.17                      & 2.93    & \textbf{1.43}                     \\ \toprule
\multicolumn{3}{c|}{Average}                & 7.52                         & 6.66                        & 5.85          & \textbf{4.66}        \\ \toprule
\end{tabular}
}
\end{table}
% \vspace{-1.0em}

\textbf{Comparison of different methods for distribution estimation.} We evaluate the performance of different methods to estimate distribution on the CASIA-SURF dataset. The results are shown in Table~\ref{ab-pe}. `PE' denotes the vanilla method that estimates the distribution for the feature vector via a fully connected layer~\cite{pe-face-1}. `PCME' further introduces attention modules to aggregate the information from the feature map to estimate the distribution for the feature vector~\cite{pe-re-1}. DMR estimates the distribution for the feature map directly via a convolutional layer. As shown in Table~\ref{ab-pe}, the proposed DMR outperforms PE and PCME by 2.00\% and 1.19\%, respectively. This demonstrates the validity of learning feature map distribution. This is because it considers more embedding parameters. More importantly, the distribution of feature maps allows DMRNet to handle dense prediction tasks, such as segmentation. 

% \textbf{Generalization to different multimodal model}

\begin{SCtable}[0.7][h]
\centering
\caption{Performance on the multimodal segmentation task with NYUv2. The metric is mIOU($\uparrow$) and the higher the value, the better the performance. $\CIRCLE$ and $\Circle$ denote available and missing modalities, respectively.}
\label{pe-se-nyu}
\begin{tabular}{|cccc}
\toprule
\multicolumn{2}{c|}{Modality}  & \multicolumn{2}{c}{Methods}                                                                      \\ \toprule
\multicolumn{1}{c}{RGB} & Depth & \multicolumn{1}{|c}{ESANet-MD} &   DMRNet     \\ \toprule
\multicolumn{1}{c}{\CIRCLE} &  $\Circle$  & \multicolumn{1}{|c}{41.34}  & \textbf{44.10} \\ 
\multicolumn{1}{c}{\Circle}  & \CIRCLE  & \multicolumn{1}{|c}{39.76}  & \textbf{41.88} \\ 
\multicolumn{1}{c}{\CIRCLE} & \CIRCLE  & \multicolumn{1}{|c}{47.23}  & \textbf{49.27} \\ \toprule
\multicolumn{2}{c}{Average}   & \multicolumn{1}{|c}{42.77}  & \textbf{45.08} \\ \toprule
\end{tabular}
\end{SCtable}
% \vspace{-1.0em}

\textbf{Generalization to dense prediction task.} We conduct multimodal segmentation experiments on the NYUv2~\cite{nyuv2} dataset to study the generalization of the proposed DMRNet to dense prediction tasks. The dataset comprises 1,449 indoor RGB-D images, of which 795 are used for training and 654 for testing. We use the common 40-class label setting and report our results on the validation set, measured by mean IOU (mIOU). 

we set ESANnet-MD as the baseline, which is the variant of ESANet~\cite{ESANet} by adding the Bernoulli indicator $\delta$ after the modality encoder. for a fair comparison, we unify the backbone and fusion module of DMRNet as those used in ESANet. While RGB-Depth semantic segmentation requires the representation after each block to predict the output, DMRNet only decouples the fusing representation at the penultimate layer to assist model optimization. Otherwise, the optimization process will not converge due to higher uncertainty introduced by multiple random representation samplings.

 % In contrast to the multimodal classification task, RGB-Depth semantic segmentation requires the representation after each block to predict the output. However, this does not mean that we need to decouple the representation after each block. DMRNet in the segmentation task also decouples the fusing representation at the penultimate layer to assist model optimization. Otherwise, the optimization process will not converge due to the uncertainty introduced by multiple random representation samplings.

As shown in Table~\ref{pe-se-nyu}, DMRNet outperforms the ESANet-MD on each modality combination consistently and improves the mIOU by 2.46\% on average.  This shows the generalization of the proposed DMRNet to enhance inference robustness for the dense prediction tasks.

\subsection{Parameters Analysis}
\label{pa}

\textbf{The effect of $\alpha$}. We first investigate the influence of different $\alpha$ for DMRNet. Here we use the decoupled multimodal representation only and $\beta$ is set as 0. The result is shown in Table~\ref{para_a}. Firstly, the model achieves the best performance when $\alpha=1e-3$. Secondly, when $\alpha$ is set as 0, the model learns a bigger variance ($\sigma^2$), i.e. 4.78, and the average ACER (8.18) is even worse than the baseline SF-MD model (7.52) that does not use decoupled multimodal representation. This demonstrates the necessity of distributional regularization for decoupled multimodal representation. Thirdly, when $\alpha$ is increasing, $\sigma^2$ will decrease, verifying the effect of $\alpha*L_{DR}$ to control the span of estimated distribution. Finally, too large $\alpha$, such as 1e-1, may cause the convergence problem since the estimated distribution becomes a completely random Gaussian distribution. Thus, for other tasks and datasets, we mainly search for the best parameter $\alpha$ among 1e-4, 1e-3, and 1e-2. And for all the CRAME-D, Kinetics-Sounds, and NYUv2 datasets, $\alpha$ is set as 1e-3.

\begin{table}[]
\centering

\begin{floatrow}
\capbtabbox{
\setlength{\tabcolsep}{1.5mm}{
\begin{tabular}{ccccccc}
\hline
\multicolumn{2}{c|}{$\alpha$}  & 0  & 1e-4 & 1e-3 & 1e-2 & 1e-1 \\ \hline
\multicolumn{2}{c|}{$\sigma^{2}$}   & 4.78 & 1.95 & 1.11 & 1.01  & 0.99 \\ \hline
\multicolumn{2}{c|}{ ACER} & 8.18 & 6.67 & \textbf{4.66} & 6.24 & 49.76 \\ \hline
\end{tabular}}
}{\caption{Results of DMRNet that leverages only the decoupled multimodal representation with different $\alpha$ on CAISA-SURF. $\sigma^2$ is the variance mean of all estimated parameters. }
\label{para_a}}
\capbtabbox{
\renewcommand\arraystretch{1.5}
\setlength{\tabcolsep}{1.0mm}{
\begin{tabular}{ccccccc}
\hline
\multicolumn{2}{c|}{$\beta$}  & 0.1 & 0.3 & 0.5 & 0.7 & 0.9 \\ \hline
% \multicolumn{2}{c|}{$L_{UDR}$}  & 0.28 & 0.19 & 0.11 & 0.09 & 0.06 \\ \hline
\multicolumn{2}{c|}{ACER} & 4.31 & 4.06 & 4.11 &  \textbf{3.58} & 4.22 \\ \hline
\end{tabular}}
}{\caption{Results of the DMRNet with different $\beta$ on CAISA-SURF. The metric is average ACER.}
\label{para_b}}

\end{floatrow}
\end{table}

% \vspace{-1.0em}

% DMRNet achieves the best performance when $\alpha$ is set as 1e-3. Secondly, when using probabilistic embedding while setting $\alpha$ as 0, the average ACER (8.18) is even worse than the baseline SF-MD model (7.52) that does not use probabilistic embedding. This demonstrates the necessity of distributional regularization for decoupled multimodal representation. Thirdly, when $\alpha$ is increasing, $L_{DR}$ will decrease and maintain $\alpha*L_{DR}$ at the same order of magnitude between 1e-1 to 1e0. This shows the effect of $\alpha$ to control the distance between the estimated distribution and the normal distribution. Finally, too large $\alpha$, such as 1e-1, may cause the converge problem since the estimated distribution become a completely random Gaussian distribution ($L_{DR}$=5e-4). Thus, for other tasks and datasets, we mainly search for the best parameter among the rest $\alpha$ 1e-4, 1e-3, and 1e-2. And for the CRAME-D and Kinetics-Sounds datasets, $\alpha$ is set as 1e-3.

% Therefore, we set $\alpha$ as 1e-3 for both multimodal classification and segmentation tasks considered in our experiments.

\textbf{The effect of $\beta$}. We then investigate the influence of different $\beta$ for DMRNet. Here $\alpha$ is set as 1e-3. The result is shown in Table~\ref{para_b}. Firstly, our method achieves the best performance when the $\beta$ is set as 0.7. Secondly, too small or large $\beta$ may lead to poor performance. On the one hand, too small $\beta$ (0.1) may cause insufficient regularization for hard combinations. On the other hand, too large $\beta$ (0.9) may limit the optimization of the primary loss $L_{TTL}$, which will also lead to sub-optimal performance. Thus, for other tasks and datasets, we mainly search for the best parameter $\beta$ among 0.3, 0.5, and 0.7. And $\beta$ for the CREMA-D, Kinetics-Sounds, and NYUV2 datasets is set as 0.3, 0.3, and 0.5, respectively.

% \begin{table}[]
% \centering
% \caption{Results of the DMRNet with different $\beta$ on CAISA-SURF. The metric is average ACER.}
% \label{para_b}
% \begin{tabular}{ccccccc}
% \hline
% \multicolumn{2}{c|}{$\beta$}  & 0.1 & 0.3 & 0.5 & 0.7 & 0.9 \\ \hline
% % \multicolumn{2}{c|}{$L_{UDR}$}  & 0.28 & 0.19 & 0.11 & 0.09 & 0.06 \\ \hline
% \multicolumn{2}{c|}{ACER} & 5.10 & 4.66 & \textbf{4.29} & 4.51 & 5.22 \\ \hline
% \end{tabular}
% \end{table}

\section{Conclusion}

In this paper, we reveal the intra-class representation constraint in conventional common subspace-based methods and propose a novel framework DMRNet to assist robust multimodal learning. It consists of two key components: decoupled multimodal representation and hard combination regularizer. Decoupled multimodal representation decouples the training and inference embedding for each sample to relax the constraint on inference embedding. Hard combination regularizer mines and regularizes the hard modality combinations to deal with the unbalanced training of different combinations. Finally, extensive experiments on multimodal classification and segmentation tasks demonstrate the effectiveness of the proposed method for incomplete multimodal learning. The code will be public after acceptance.

\bibliographystyle{splncs04}
\bibliography{main}

\begin{thebibliography}{10}
\providecommand{\url}[1]{\texttt{#1}}
\providecommand{\urlprefix}{URL }
\providecommand{\doi}[1]{https://doi.org/#1}

\bibitem{mem1}
Arpit, D., Jastrz{\k{e}}bski, S., Ballas, N., Krueger, D., Bengio, E., Kanwal, M.S., Maharaj, T., Fischer, A., Courville, A., Bengio, Y., et~al.: A closer look at memorization in deep networks. In: International conference on machine learning. pp. 233--242. PMLR (2017)

\bibitem{special-hall1}
C, J., Stroud, D.A., Ross, C., Sun, J., Deng, R., Sukthankar: D3d: Distilled 3d networks for video action recognition. In: Proceedings of the IEEE/CVF Winter Conference on Applications of Computer Vision. pp. 2755--2764 (2020)

\bibitem{MC3}
Cai, L., Wang, Z., Gao, H., Shen, D., Ji, S.: Deep adversarial learning for multi-modality missing data completion. In: Proceedings of the ACM SIGKDD International Conference on Knowledge Discovery \& Data Mining. pp. 1158--1166 (2018)

\bibitem{rgbd_seg1}
Cao, J., Leng, H., Lischinski, D., Cohen-Or, D., Tu, C., Li, Y.: Shapeconv: Shape-aware convolutional layer for indoor rgb-d semantic segmentation. In: Proceedings of the IEEE/CVF International Conference on Computer Vision. pp. 7088--7097 (2021)

\bibitem{pe-re-1}
Chun, S., Oh, S.J., De~Rezende, R.S., Kalantidis, Y., Larlus, D.: Probabilistic embeddings for cross-modal retrieval. In: Proceedings of the IEEE/CVF Conference on Computer Vision and Pattern Recognition. pp. 8415--8424 (2021)

\bibitem{special-hall5}
Crasto, N., Weinzaepfel, P., Alahari, K., Schmid, C.: Mars: Motion-augmented rgb stream for action recognition. In: Proceedings of the IEEE/CVF Conference on Computer Vision and Pattern Recognition. pp. 7882--7891 (2019)

\bibitem{rfnet}
Ding, Y., Yu, X., Yang, Y.: Rfnet: Region-aware fusion network for incomplete multi-modal brain tumor segmentation. In: Proceedings of the IEEE/CVF International Conference on Computer Vision. pp. 3975--3984 (2021)

\bibitem{pe-face-2}
.et.al, C.J.: Data uncertainty learning in face recognition. In: CVPR2020

\bibitem{special-hall4}
Garcia, N.C., Morerio, P., Murino, V.: Modality distillation with multiple stream networks for action recognition. In: Proceedings of the European Conference on Computer Vision. pp. 103--118 (2018)

\bibitem{special-hall6}
Garcia, N.C., Morerio, P., Murino, V.: Learning with privileged information via adversarial discriminative modality distillation. IEEE transactions on pattern analysis and machine intelligence  \textbf{42}(10),  2581--2593 (2019)

\bibitem{mem2}
Han, B., Yao, Q., Yu, X., Niu, G., Xu, M., Hu, W., Tsang, I., Sugiyama, M.: Co-teaching: Robust training of deep neural networks with extremely noisy labels. Advances in neural information processing systems  \textbf{31} (2018)

\bibitem{channel}
Han, K., Wang, Y., Tian, Q., Guo, J., Xu, C., Xu, C.: Ghostnet: More features from cheap operations. In: Proceedings of the IEEE/CVF conference on computer vision and pattern recognition. pp. 1580--1589 (2020)

\bibitem{hemis}
Havaei, M., Guizard, N., Chapados, N., Bengio, Y.: Hemis: Hetero-modal image segmentation. In: International Conference on Medical Image Computing and Computer-Assisted Intervention. pp. 469--477. Springer (2016)

\bibitem{special-hall3}
Hoffman, J., Gupta, S., Darrell, T.: Learning with side information through modality hallucination. In: Proceedings of the IEEE Conference on Computer Vision and Pattern Recognition. pp. 826--834 (2016)

\bibitem{mm_cf2}
Hong, D., Hu, J., Yao, J., Chanussot, J., Zhu, X.X.: Multimodal remote sensing benchmark datasets for land cover classification with a shared and specific feature learning model. ISPRS Journal of Photogrammetry and Remote Sensing  \textbf{178},  68--80 (2021)

\bibitem{rgbd_seg2}
Hu, X., Yang, K., Fei, L., Wang, K.: Acnet: Attention based network to exploit complementary features for rgbd semantic segmentation. In: 2019 IEEE International Conference on Image Processing (ICIP). pp. 1440--1444. IEEE (2019)

\bibitem{gan-reprentation-2}
Jeong, S.w., Cho, H.h., Lee, S., Park, H.: Robust multimodal fusion network using adversarial learning for brain tumor grading. Computer Methods and Programs in Biomedicine  \textbf{226},  107165 (2022)

\bibitem{mm_detection3}
Jin, W.D., Xu, J., Han, Q., Zhang, Y., Cheng, M.M.: Cdnet: Complementary depth network for rgb-d salient object detection. IEEE Transactions on Image Processing  \textbf{30},  3376--3390 (2021)

\bibitem{MC2}
Jue, J., Jason, H., Neelam, T., Andreas, R., Sean, B.L., Joseph, D.O., Harini, V.: Integrating cross-modality hallucinated mri with ct to aid mediastinal lung tumor segmentation. In: International Conference on Medical Image Computing and Computer-Assisted Intervention. pp. 221--229. Springer (2019)

\bibitem{rep}
Kingma, D.P., Welling, M.: Auto-encoding variational bayes. arXiv preprint arXiv:1312.6114  (2013)

\bibitem{pe-word-3}
Li, X., Vilnis, L., Zhang, D., Boratko, M., McCallum, A.: Smoothing the geometry of probabilistic box embeddings. In: International Conference on Learning Representations (2019)

\bibitem{special-hall2}
Li, X., Lei, L., Sun, Y., Kuang, G.: Dynamic-hierarchical attention distillation with synergetic instance selection for land cover classification using missing heterogeneity images. IEEE Transactions on Geoscience and Remote Sensing  \textbf{60},  1--16 (2021)

\bibitem{gan-reprentation-1}
Lin, Y., Gou, Y., Liu, Z., Li, B., Lv, J., Peng, X.: Completer: Incomplete multi-view clustering via contrastive prediction. In: Proceedings of the IEEE/CVF conference on computer vision and pattern recognition. pp. 11174--11183 (2021)

\bibitem{spcical-gan1}
Liu, A., Tan, Z., Wan, J., Liang, Y., Lei, Z., Guo, G., Li, S.Z.: Face anti-spoofing via adversarial cross-modality translation. IEEE Transactions on Information Forensics and Security  \textbf{16},  2759--2772 (2021)

\bibitem{mm_cf1}
Liu, A., Wan, J., Escalera, S., Jair~Escalante, H., Tan, Z., Yuan, Q., Wang, K., Lin, C., Guo, G., Guyon, I., et~al.: Multi-modal face anti-spoofing attack detection challenge at cvpr2019. In: Proceedings of the IEEE/CVF Conference on Computer Vision and Pattern Recognition Workshops. pp.~0--0 (2019)

\bibitem{gan-issue1}
Liu, H., Ma, S., Xia, D., Li, S.: Sfanet: A spectrum-aware feature augmentation network for visible-infrared person re-identification. arXiv preprint arXiv:2102.12137  (2021)

\bibitem{pe-re-2}
Neculai, A., Chen, Y., Akata, Z.: Probabilistic compositional embeddings for multimodal image retrieval. In: Proceedings of the IEEE/CVF Conference on Computer Vision and Pattern Recognition. pp. 4547--4557 (2022)

\bibitem{pe-word-2}
Neelakantan, A., Shankar, J., Passos, A., McCallum, A.: Efficient non-parametric estimation of multiple embeddings per word in vector space. arXiv preprint arXiv:1504.06654  (2015)

\bibitem{ogm}
Peng, X.e.: Balanced multimodal learning via on-the-fly gradient modulation. In: CVPR2022

\bibitem{device1}
Pinto, A., Pedrini, H., Schwartz, W.R., Rocha, A.: Face spoofing detection through visual codebooks of spectral temporal cubes. IEEE Transactions on Image Processing  \textbf{24}(12),  4726--4740 (2015)

\bibitem{rgbd_seg3}
Seichter, D., K{\"o}hler, M., Lewandowski, B., Wengefeld, T., Gross, H.M.: Efficient rgb-d semantic segmentation for indoor scene analysis. In: 2021 IEEE International Conference on Robotics and Automation (ICRA). pp. 13525--13531. IEEE (2021)

\bibitem{ESANet}
Seichter, D., K{\"o}hler, M., Lewandowski, B., Wengefeld, T., Gross, H.M.: Efficient rgb-d semantic segmentation for indoor scene analysis. In: 2021 IEEE International Conference on Robotics and Automation (ICRA). pp. 13525--13531. IEEE (2021)

\bibitem{pe-face-1}
Shi, Y., Jain, A.K.: Probabilistic face embeddings. In: Proceedings of the IEEE/CVF International Conference on Computer Vision. pp. 6902--6911 (2019)

\bibitem{nyuv2}
Silberman, N., Hoiem, D., Kohli, P., Fergus, R.: Indoor segmentation and support inference from rgbd images. In: European conference on computer vision. pp. 746--760. Springer (2012)

\bibitem{pe-pose}
Sun, J.J., Zhao, J., Chen, L.C., Schroff, F., Adam, H., Liu, T.: View-invariant probabilistic embedding for human pose. In: Computer Vision--ECCV 2020: 16th European Conference, Glasgow, UK, August 23--28, 2020, Proceedings, Part V 16. pp. 53--70. Springer (2020)

\bibitem{mm_detection2}
Sun, P., Zhang, W., Wang, H., Li, S., Li, X.: Deep rgb-d saliency detection with depth-sensitive attention and automatic multi-modal fusion. In: Proceedings of the IEEE/CVF conference on computer vision and pattern recognition. pp. 1407--1417 (2021)

\bibitem{mm_cf3}
Tian, H., Tao, Y., Pouyanfar, S., Chen, S.C., Shyu, M.L.: Multimodal deep representation learning for video classification. World Wide Web  \textbf{22}(3),  1325--1341 (2019)

\bibitem{pe-word-1}
Vilnis, L., McCallum, A.: Word representations via gaussian embedding. arXiv preprint arXiv:1412.6623  (2014)

\bibitem{shaspec}
Wang, H., Chen, Y., Ma, C., Avery, J., Hull, L., Carneiro, G.: Multi-modal learning with missing modality via shared-specific feature modelling. In: Proceedings of the IEEE/CVF Conference on Computer Vision and Pattern Recognition. pp. 15878--15887 (2023)

\bibitem{mmanet}
Wei, S., Luo, C., Luo, Y.: Mmanet: Margin-aware distillation and modality-aware regularization for incomplete multimodal learning. In: Proceedings of the IEEE/CVF Conference on Computer Vision and Pattern Recognition. pp. 20039--20049 (2023)

\bibitem{mem3}
Zhang, C., Bengio, S., Hardt, M., Recht, B., Vinyals, O.: Understanding deep learning (still) requires rethinking generalization. Communications of the ACM  \textbf{64}(3),  107--115 (2021)

\bibitem{surf}
Zhang, S., Wang, X., Liu, A., Zhao, C., Wan, J., Escalera, S., Shi, H., Wang, Z., Li, S.Z.: A dataset and benchmark for large-scale multi-modal face anti-spoofing. In: Proceedings of the IEEE/CVF Conference on Computer Vision and Pattern Recognition. pp. 919--928 (2019)

\bibitem{mmformer}
Zhang, Y., He, N., Yang, J., Li, Y., Wei, D., Huang, Y., Zhang, Y., He, Z., Zheng, Y.: mmformer: Multimodal medical transformer for incomplete multimodal learning of brain tumor segmentation. arXiv preprint arXiv:2206.02425  (2022)

\bibitem{mm_detection1}
Zhou, T., Fan, D.P., Cheng, M.M., Shen, J., Shao, L.: Rgb-d salient object detection: A survey. Computational Visual Media  \textbf{7}(1),  37--69 (2021)

\bibitem{lcr}
Zhou, T., Canu, S., Vera, P., Ruan, S.: Brain tumor segmentation with missing modalities via latent multi-source correlation representation. In: International Conference on Medical Image Computing and Computer-Assisted Intervention. pp. 533--541. Springer (2020)

\bibitem{gan-issue2}
Zhu, P., Yao, X., Wang, Y., Cao, M., Hui, B., Zhao, S., Hu, Q.: Latent heterogeneous graph network for incomplete multi-view learning. IEEE Transactions on Multimedia  (2022)

\end{thebibliography}
\end{document}